\documentclass[conference]{IEEEtran}
\IEEEoverridecommandlockouts
\usepackage{cite}
\usepackage{amsmath,amssymb,amsfonts}
\usepackage{algorithmic}
\usepackage{graphicx}
\usepackage[bookmarks=false]{hyperref}
\usepackage{textcomp}
\usepackage{xcolor}
\usepackage{subfig}
\hypersetup{
    colorlinks=true,
    linkcolor=blue,urlcolor=blue}
\def\BibTeX{{\rm B\kern-.05em{\sc i\kern-.025em b}\kern-.08em
    T\kern-.1667em\lower.7ex\hbox{E}\kern-.125emX}}
\begin{document}

\title{FigureNet : A Deep Learning model for Question-Answering on Scientific Plots
}


\author{\IEEEauthorblockN{Revanth Reddy}
\IEEEauthorblockA{\textit{Computer Science and Engineering} \\
\textit{Indian Institute of Technology Madras}\\
Chennai, India \\
g.revanthreddy111@gmail.com}
\and
\IEEEauthorblockN{Rahul Ramesh}
\IEEEauthorblockA{\textit{Computer Science and Engineering} \\
\textit{Indian Institute of Technology Madras}\\
Chennai, India \\
rahul13ramesh@gmail.com} 
\and
\IEEEauthorblockN{Ameet Deshpande}
\IEEEauthorblockA{\textit{Computer Science and Engineering} \\
\textit{Indian Institute of Technology Madras}\\
Chennai, India \\
ameetsd97@gmail.com}
\and
\IEEEauthorblockN{Mitesh M. Khapra}
\hspace{48em}
\IEEEauthorblockA{\textit{Computer Science and Engineering} \\
\textit{Indian Institute of Technology Madras}\\
Chennai, India \\
miteshk@cse.iitm.ac.in}
}

\maketitle

\begin{abstract}
Deep Learning has managed to push boundaries in a wide variety of tasks. One area of interest is to tackle problems in reasoning and understanding, with an aim to emulate human intelligence.  In this work, we describe a deep learning model that addresses the reasoning task of question-answering on categorical plots. We introduce a novel architecture FigureNet, that learns to identify various plot elements, quantify the represented values and determine a relative ordering of these statistical values. We test our model on the FigureQA dataset which provides images and accompanying questions for scientific plots like bar graphs and pie charts, augmented with rich annotations. Our approach outperforms the state-of-the-art Relation Networks baseline by approximately $7\%$ on this dataset, with a training time that is over an order of magnitude lesser.  
\end{abstract}


\section{Introduction}
Deep learning has transformed the computer vision and natural language processing landscapes and has become a ubiquitous tool in their associated applications. The potential of convolutional neural networks on images was demonstrated with its success in the ImageNet classification task \cite{krizhevsky2012imagenet}. Long-short-term Memory networks \cite{hochreiter1997long} have demonstrated a capability to tackle complex tasks like sentence summarization \cite{rush2015neural}, machine passage comprehension \cite{ReadComprehend} and Neural Machine translation \cite{bahdanau2014neural}. 
Neural network models are in-fact a result of preliminary attempts to model the brain and hence it is a natural area of interest to accurately model ``reasoning". A plethora of visual reasoning tasks \cite{lin2014microsoft,johnson2017clevr} have been created to benchmark these capabilities of neural networks.  Visual question answering tasks require a combination of reasoning, Natural Language Processing and Computer Vision techniques. The model must be capable of obtaining representations of the image and question apart from intelligently combining these representations to generate an answer. This task helps machines gain the ability to process visual signals and use it to solve multi-modal problems.

The rudimentary Convolutional neural networks (CNN) and Long-short term memory networks (LSTM) models are incapable of handling these datasets.  \cite{haehn2018evaluating} have demonstrated that CNNs aren't a satisfactory model for human graphical perception and fail when applied to data visualizations. However, reasoning specific architectures have managed to achieve super-human scores on these reasoning based tasks \cite{perez2017film,santoro2017simple}. One point of note is that these datasets have predominantly addressed spatial and relational reasoning. \cite{kahou2017figureqa} designed a dataset that uses scientific graphs and figures to test count-based, numeric, spatial and relational reasoning. Scientific figures are a compact representation of statistical information. They are found not only in scientific research papers but also in business analysis reports, consensus reports and various other sources wherein it is possible to supplement textual information with figures. Therefore, automating the  understanding of this visual information could aid human analysts since it allows drawing inferences from various reports and papers. An architecture addressing this task is hence of great utility since it bridges the gap towards a universal reasoning module.

We propose a neural network architecture \textit{FigureNet}, that incorporates various entities in scientific plots, to address the reasoning task. FigureNet is motivated by the principle of divide and conquer. Different modules are used to emulate different logical components and are put together, while also ensuring that the model is end-to-end differentiable. In order to ensure that the functionality of the modules are made clear, we employ supervised-pretraining on each of the modules on relevant individual sub-tasks.

We compare our model against the Relation Networks (RN) architecture \cite{santoro2017simple} and a standard CNN-LSTM architecture. We evaluate the efficacy of our model using the categorical plots present in the Figure-QA dataset. Our model outperforms these baselines with a training time that is over an order of magnitude lesser than that of Relation networks. The rest of the paper is structured as follows. Section 2 gives the related work for this paper. Section 3 describes the FigureQA dataset \cite{kahou2017figureqa} and introduces the RN baseline. In Section 4, we lay out our approach for question answering on categorical plots like bar graphs and pie charts. In Section 5, we explain our training process and show improvements over various baselines on the FigureQA dataset. This is followed by a collection of ablation studies that dissect the different components of our model. Section 6 gives a methodology for extending our approach to real-life figures. Finally, Section 7 concludes the paper and highlights directions for future work.

\section{Related work}
There are a variety of visual question-answering datasets \cite{lin2014microsoft,johnson2017clevr}. These datasets however have questions which solely deal with the positional relationship between objects. Hence, the main function of the neural network here is to identify different objects and codify their positions. 

The baselines for this task involve naively combining the LSTM and CNN architectures. \cite{ren2015exploring} describe an end-to-end differentiable architecture which sets the bar for neural networks on spatial reasoning tasks. \cite{malinowski2017ask} report results on a varied set of combinations of textual model embeddings and image embeddings. These baselines were consequently superceded by attention based models which use the image embeddings to generate attention maps over the text \cite{nam2016dual}. Parallel to the development of attention based architectures, several pieces of work in literature explored different fusion functions that combine image and sentence representations \cite{ben2017mutan}. A large body of work also addresses the Visual question-answering problem using modular networks wherein different modules are used to replicate different logical components \cite{andreas2016neural,hu2017learning}.  The state of the art approaches in visual question answering use a rather simple, end-to-end differentiable model and achieve super-human performance on relational reasoning tasks \cite{perez2017film,santoro2017simple}. 

There is a plethora of literature on the advantages of pre-training in deep learning. \cite{erhan2009difficulty} discuss the difficulty of training deep architectures and the effect of unsupervised pre-training. They infer that starting the supervised optimization from pre-trained weights rather than from random initialized weights consistently yields better performing classifiers. \cite{erhan2010does} suggest that unsupervised pre-training acts as a regularizer and guides the learning towards basins of attraction of minima that support better generalization from the training data set. 

The disadvantage of Relation Networks, FiLM \cite{perez2017film}  is the computational demand of these models. Our architecture is computationally lightweight in comparison. A key requirement for our neural network model is to identify colours. Traditional convolutional layers typically mix the information content present in various channels. Inspired by the depth-wise separate convolution operation present in the Xception model \cite{chollet2016xception}, we adopt a similar family of convolution models in our design.

\section{Preliminaries}
In this section, we first describe the FigureQA dataset\footnote[1]{\url{https://datasets.maluuba.com/FigureQA}} which was introduced by \cite{kahou2017figureqa}. This is followed by a description of the Relation Networks baseline for this dataset. We consider Relation Networks as our baseline since \cite{santoro2017simple} have shown that they outperform FiLM\cite{perez2017film} on relational reasoning.

\subsection{The FigureQA Dataset}

FigureQA \cite{kahou2017figureqa} is a visual reasoning corpus which contains over a million question-answer pairs which are grounded in scientific style figures. This synthetic corpus has been designed to focus specifically on reasoning. FigureQA also has the advantage of not requiring text identification modules like OCR, since plot elements are colour-coded. It follows the general Visual Question answering setup, but also provides annotated data with bounding boxes for each figure.

\begin{figure}[htb]
	\centering
	\includegraphics[scale=0.26]{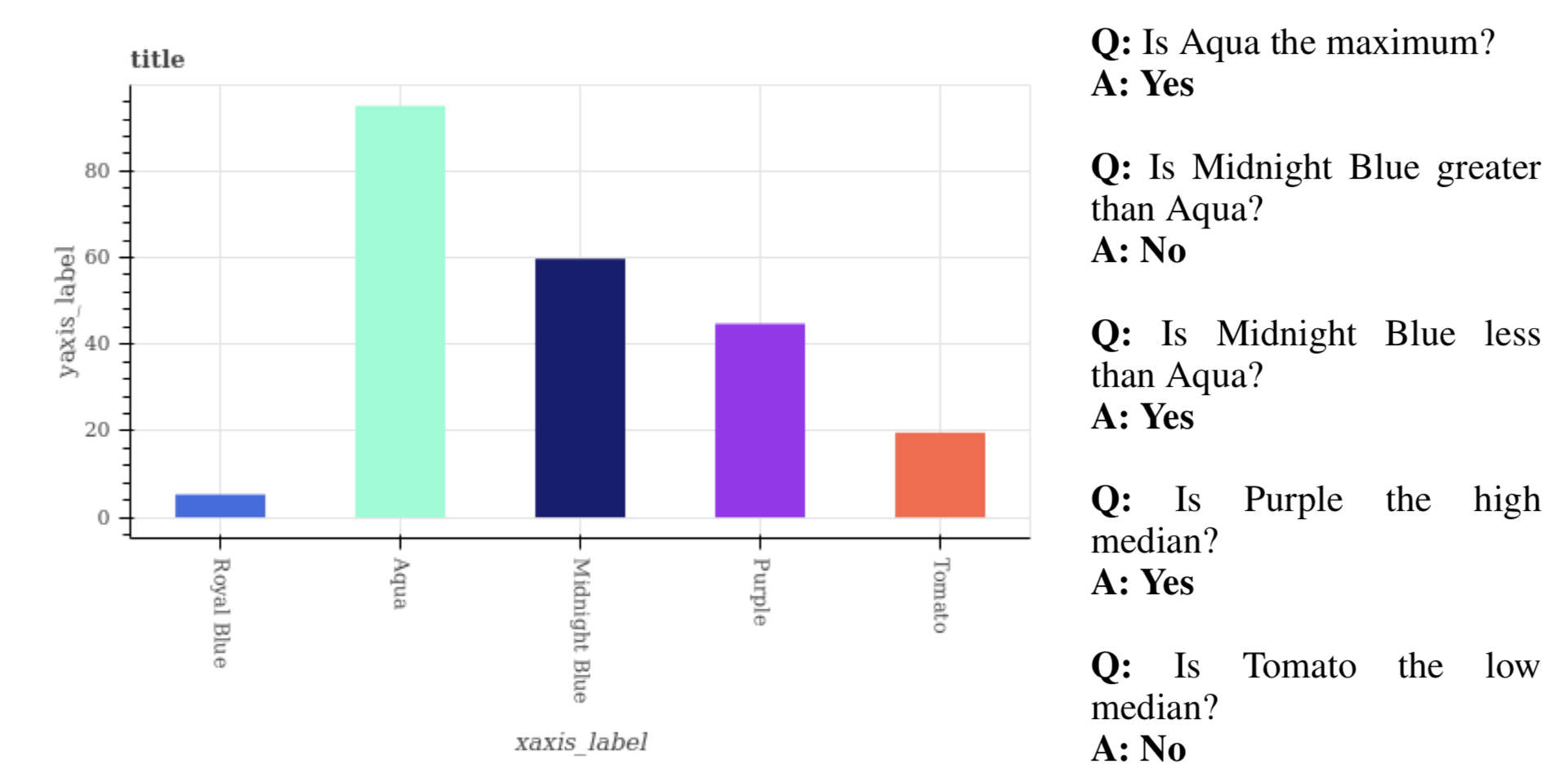}
	\caption{Vertical Bar graph with question-answer pairs}
	\label{fig:fig1}
\end{figure}

\begin{figure}[htb]
  \centering
  \subfloat{\includegraphics[scale=0.55]{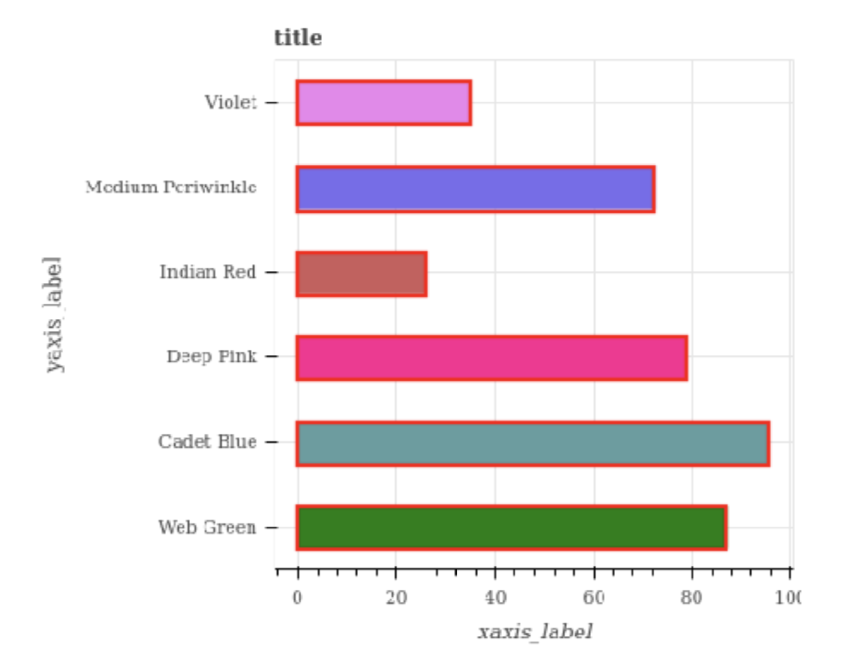}}\quad
  \subfloat{\includegraphics[scale=0.55]{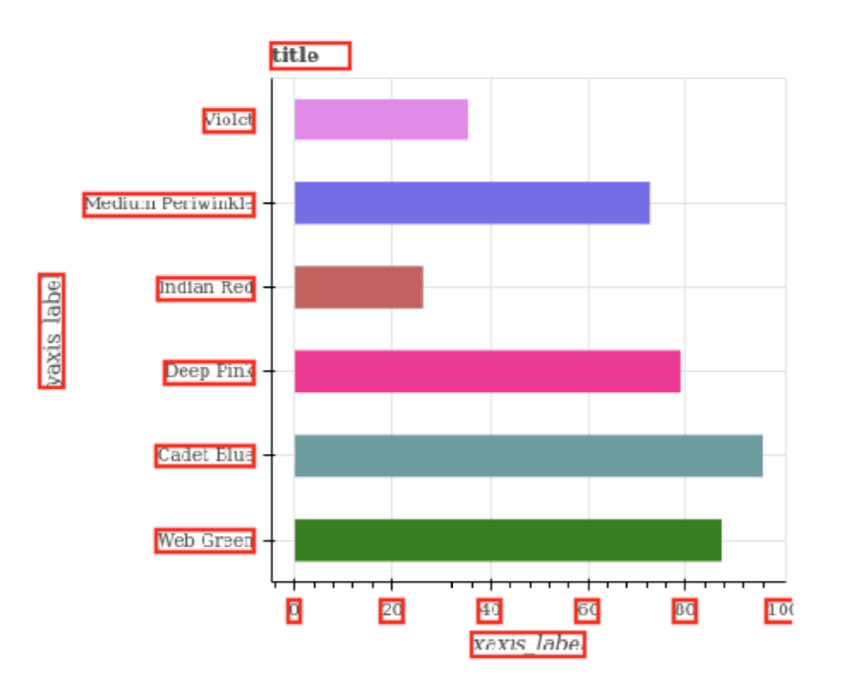}}
  \caption{Horizontal Bar graph with annotations}
  \label{fig:2}
\end{figure}

\begin{figure}[htb]
	\centering
	\includegraphics[scale=0.27]{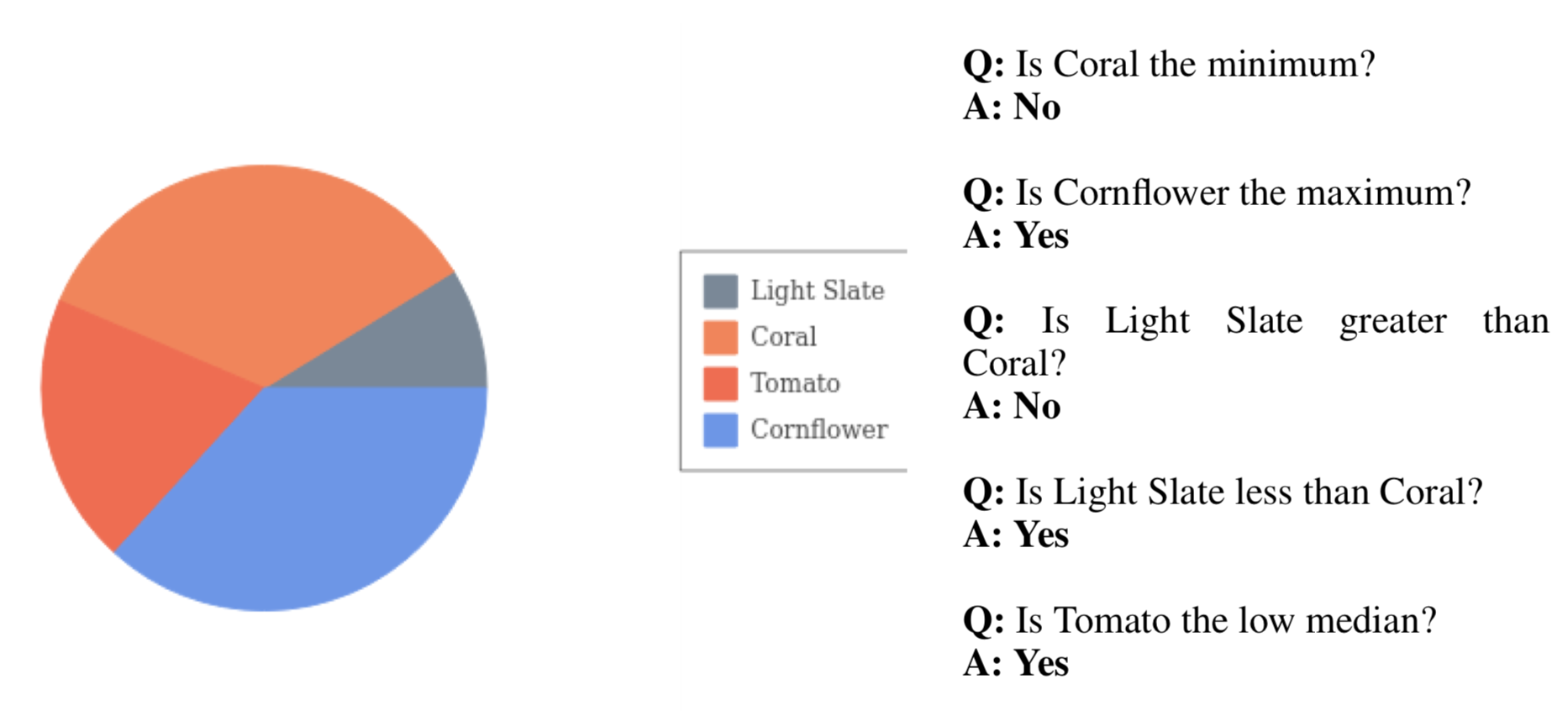}
	\caption{Pie chart with question-answer pairs}
	\label{fig:fig4}
\end{figure}

100 unique colours covering the entire spectrum of colours, were chosen from the X11 named colour set. FigureQA's training, validation and test sets are constructed such that all 100 colours are seen during training. 
Figure \ref{fig:fig1} and Figure \ref{fig:fig4} are examples of different figure types with question-answer pairs. Figure \ref{fig:2} shows an example for annotations available for each figure. Images taken from \cite{kahou2017figureqa}.

\subsection{Relation Networks}

Relation networks(RN) were introduced by \cite{santoro2017simple} as a simple yet powerful neural module for relational reasoning. Relation Networks have the ability to compute relations, just as convolutional neural networks have the ability to generate image feature map and recurrent neural networks have the ability to capture sequential dependencies. RNs have been demonstrated to achieve  a state-of-the-art, superhuman performance on a challenging dataset called CLEVR \cite{johnson2017clevr}. RN takes the object representation as input and processes the relations between objects as follows:
\begin{equation}
 RN(O) = f_{\phi}\left(\frac{1}{N^2}\sum_{i,j}g_{\theta}(o_{i,.},o_{j,.})\right)
 \end{equation}
 where $O\in R^{N\times C}$ is a matrix in which the $i^{th}$ row contains the object representation $o_{i,.}$. Here, $g_{\theta}$ calculates the relations between a pair of objects and $f_{\phi}$ aggregates these relations and computes the final output of the model. 
 
 For the FigureQA evaluation, the object representations are obtained from a convolutional neural network. The CNN output contains 64 feature maps each of size $8\times8$. Each pixel from this output corresponds to an object $o_{i,.}$. We have 64($8 \times 8$) such objects wherein each object has a 64 dimensional representation. The row and column coordinates of the pixel are appended to the corresponding object's representation so as to include the information about location of objects inside the feature map.
\begin{equation}
o_i = \left(o_{i,1}, \cdots, o_{i,64},\left\lfloor \frac{i-1}{8} \right\rfloor,(i-1) mod 8 \ \right)
\end{equation}
The input to the relation network is the set of all pairs of object representations, which are concatenated with the question encoding. The question encoding is obtained from an LSTM which has a hidden unit size of 256 in the RN baseline. $g_{\theta}$ processes each of the object pairs separately to produce a representation for the relations between the objects. These relation representations are then summed up and given as input to $f_{\phi}$, which gives the final output. For training the model, four parallel workers were used. The average of the gradients from the workers was used to update the parameters. 



\section{FigureNet}
In this section, we describe the FigureNet\footnote{Code is available at \href{https://github.com/revanth1996/FigureNet}{FigureNet}} architecture that tackles the question-answering task on bar plots and pie charts. These plots have bars or sectors present in them, which we refer to as plot elements. In these figure types, the plot elements are generally distinguished by their respective colours. Thus, we can recognize a plot element by identifying the colour in which it is drawn. For example, in Figure \ref{fig:fig1}, we can see that the five vertical bars are drawn in five different colours. Each image represents a sequence of numeric values and obtaining this sequence allows one to answer \textit{any} relevant question. For the FigureQA dataset in particular, the absolute values are not required and the relative ordering suffices. For example, in Figure \ref{fig:fig1}, the relative ordering of the five bars is [1,5,4,3,2]. The lower numbers represent lower numerical values for plot elements and this representation allows questions involving maximum, greater than, high median etc.. to be answered easily.

We hypothesize that tackling the larger task of answering the questions can be solved by handling the subtasks of identifying plot elements followed by arriving at a relative ordering of plot elements. We employ supervised pre-training for each of the subtasks, using the annotations provided in FigureQA dataset. The model is comprised of modules which are logically intended to tackle one specific subtask each. 

\subsection{Spectral Segregator Module}


\begin{figure*}[th]
    \centering
    \subfloat{\includegraphics[width=0.99\textwidth]{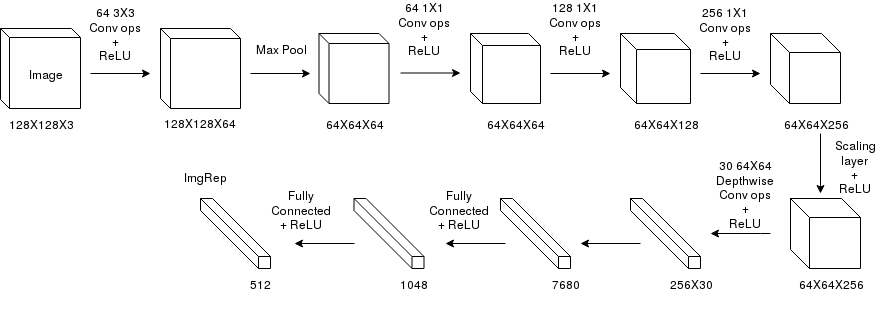}}
	\caption{Architecture of Spectral Segregator Module - Image visualizes the sequence of convolution operations}
	\label{fig:fig6}
\end{figure*}

The purpose of this module is to identify all plot elements and the colour of each of these elements. For vertical bar graphs, the model identifies the plot elements from left to right, for horizontal bar graphs, from bottom to top and for pie charts, in an anti-clockwise direction(starting from 0 degrees). The module takes the figure as input and outputs the probabilities of colours for each of the plot elements. By taking advantage of the fact that the number of plot elements in bar graphs and pie charts of FigureQA is always less than 11, the module has 11 output units where each output unit is a probability distribution over the 100 colours. For example, in Figure \ref{fig:fig1}, the targets for the module would be [Royal Blue, Aqua, Midnight Blue, Purple, Tomato, STOP, STOP, STOP, STOP, STOP, STOP] where \textit{STOP} represents that there are no more plot elements present and \textit{Royal Blue} represents a one-hot vector(probability distribution with a unit probability for the colour Royal Blue).

Traditional convolution layers do not suffice since they tend to aggregate the information and give an activation map that is a coarse representation of the image. Another peculiarity of the convolution operation is that the information across channels are summed over. Ideally, the channel information is required to be separated. Hence we solely use $1\times1$ convolutions followed by scaling layers and depthwise convolutions.

The input to this module is an image with dimensions $128\times 128\times3$. The first convolutional layer filters the input image with 64 kernels of size $3\times3\times3$. This is followed by a max-pooling layer that lowers the 2D feature map dimensions to $64\times64$. The second, third and fourth convolution layers apply $1\times1$ convolutions with number of filters for each layer being 64, 128 and 256 respectively. The output feature map is of dimensions $64\times64\times256$. This is followed by a scaling layer that performs channel-wise multiplication of each of the 256 channels. In other words, each channel $c$ is multiplied by a scalar parameter $p_c$. This operation will not change the dimensions of the feature map. The idea behind adding the $1\times1$ convolution layers and scaling layer is that different colours have different channel values and these operations will help differentiate between the colours.

In the next layer, we perform depthwise convolutions with 30 kernels of size $64\times64$ each. Since there are 256 channels in the feature map, each kernel will produce a 256 dimensional vector, thereby giving an output with dimensions $30\times256$. We add two fully connected layers on top of this, with 1048 and 512 hidden units respectively to finally output a 512 dimensional image representation. The motivation behind adding the depthwise convolutions is that each $64\times64$ filter can be understood to aggregate the count of a particular colour, thereby quantifying the values represented by various coloured plot elements. These convolution operations are visualized in \ref{fig:fig6}.

\begin{figure}[th]
    \centering
    \subfloat{\includegraphics[scale=0.31]{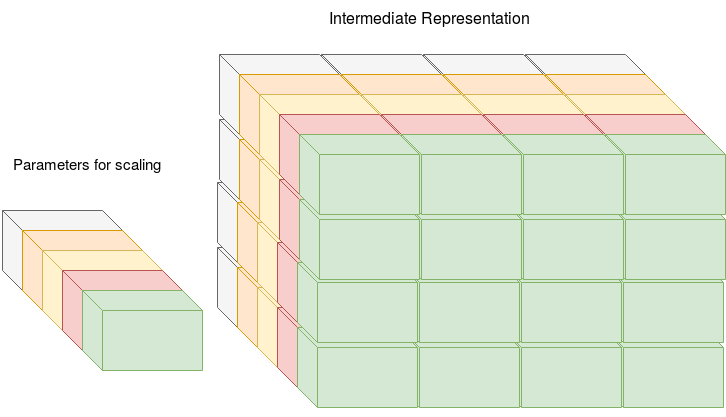}}
	\caption{Scaling layer : Each channel is multiplied by a parameter $p_c$ with each channel assigned a distinct colour}
	\label{fig:fig63}
\end{figure}

Regular convolutions group information across the channels. However, aggregating this information is counter-productive to the task of identifying and segregating the colours. Convolution layers are not designed to perfectly identify colours since the activation functions are often one-sided. Hence, the depthwise convolutions and scaling operations facilitate this requirement and equip the model, with an ability to differentiate colours. Figure \ref{fig:fig63} is a visualization of the scaling operation and the depthwise convolution is a standard 3D convolution operation. 

\begin{figure}[th]
    \centering
    \subfloat{\includegraphics[scale=0.39]{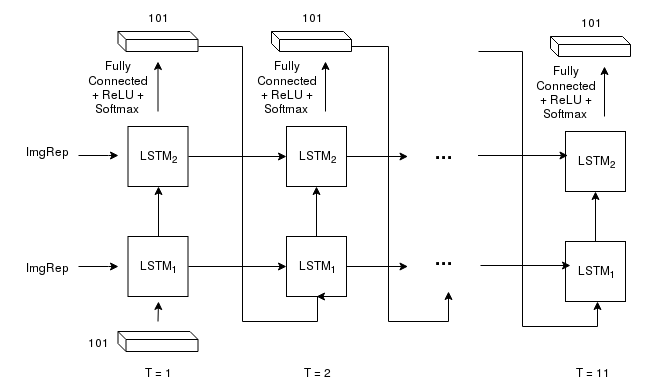}}
	\caption{Architecture of Spectral Segregator Module - Custom LSTM architecture}
	\label{fig:fig62}
\end{figure}

Finally, to output the colour probabilities for each plot element, we use a modified version of a two layered LSTM network. The architecture for this can be seen in Figure \ref{fig:fig62}. The 512 dimensional image representation is the initial state that is input to the LSTM. The output at every time-step is a probability distribution over the 100 colours and STOP label. Output at time step $t$ gives the probability of colours for the $t^{th}$ plot element. In order to mitigate the differences between the training and testing phases, the output probabilities at time step $t-1$ are given as input to the LSTM at time step $t$. This is different from a traditional LSTM in which the output is sampled from the probabilities at time step $t-1$ and then given as input at time step $t$, i.e we do away with the sampling. This also allows propagating gradients from input at time step $t$ to the output of time-step $t-1$. The input at time step 1 is a 101 dimensional parameter that is learned by the network. The motivation behind using an LSTM mainly comes from the fact that the number of plot elements in a figure is not fixed and we found that using an LSTM performs better than predicting the 11 outputs at one go. If $h^1_{t-1},s^1_{t-1}$ and $h^2_{t-1},s^2_{t-1}$  are hidden states at time step t-1 for first layer and second layer respectively, the equations for finding the output probabilities at time step t are given below:
\begin{equation}
h^1_t,s^1_t = LSTM_1(o_{t-1},h^1_{t-1},s^1_{t-1})
\end{equation}
\begin{equation}
h^2_t,s^2_t = LSTM_2([h^1_t,s^1_t],h^2_{t-1},s^2_{t-1})
\end{equation}
\begin{equation}
o_t = softmax(ReLU(W^Th^2_t + b))
\end{equation}

\subsection{Order Extraction Module}

This module identifies and quantifies the statistical values of each plot element, followed by sorting these values into a linear order. Since the number of plot elements in bar graphs and pie charts of FigureQA is always less than 11, the possible positions in the sorted order are [1,2,3,4,5,6,7,8,9,10], where lower numbers represent lower statistical values, with 0 being reserved as order for plot elements that are absent. For example, in Figure \ref{fig:fig1}, the targets for the Order Extraction module would be one-hot values of [1,5,4,3,2,0,0,0,0,0,0](i.e each element is one-hot vector). The module takes the image as input and gives the probabilities for the position in the sorted order of each of the plot element as output. We observed that the final feed-forward network learns to ignore the output probabilities for the plot elements which are absent.

The architecture for this module is almost the same as that of the Spectral Segregator module except that it has three fully connected layers with 2048, 1024, 512 hidden units respectively, after the depthwise convolutions. The output of two layered LSTM network at each time step is a probability distribution over the 11 possible relative ordering values(0 to 10). The additional parameters are required to perform the heavy lifting of the sorting operation.

\subsection{Final Feed-forward network}

\begin{figure*}[th]
	\centering
	\includegraphics[scale=0.41]{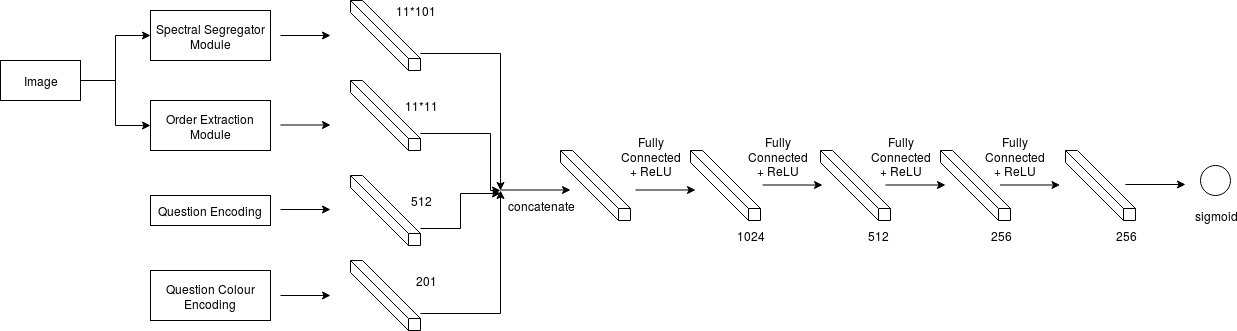}
	\caption{Architecture of final feedforward network}
	\label{fig:fig7}
\end{figure*}

We concatenate the output probabilities from the 11 timesteps in the Spectral Segregation and Order Extraction modules. Thus, we get a $11\times101 + 11\times11 = 1232$ dimensional figure representation. We consider the output probabilities instead of sampling the outputs so that we can backpropagate the gradients through these modules when the entire network is trained end-to-end. The question representation consists of two parts, question encoding and question-colour encoding. The question encoding is produced by an LSTM with 256 hidden units. The question-colour encoding is a representation of the colours that are present in the question. It is obtained by concatenating the 100 dimensional one-hot vector of first colour in question with the 101 dimensional(100 colours + one label for no second colour) one-hot vector of second colour in question. The question encoding and question-colour encoding together form the question representation. The figure representation is concatenated with question representation and given as input to feed-forward neural network.

The feed-forward network has four hidden layers and one output layer. The hidden layers have 1024, 512, 256 and 256 hidden units respectively and the output layer has only 1 unit. The activations are ReLU for the hidden layers and sigmoid for the output layer. The architecture is shown in the Figure \ref{fig:fig7}.

\section{Experiments}
The training set contains 60,000 images with 20,000 each for vertical bar graphs, horizontal bar graphs and pie charts. The validation and test sets contain 12,000 images each with an equal split for the 3 figure types. The annotations are available for the training and validation set but not for the test set.
For the supervised pre-training task, the targets for the modules are generated from the annotations for each image provided in the FigureQA dataset. Note that the annotations are used only while pre-training the modules. The final end-to-end model answers the questions by taking only the image as input.

\subsection{Training Specifics}

For pre-training the modules, a cross entropy loss between the softmax output probabilities at each time step and the one-hot targets generated from the annotations, are utilized. For the question answering task, a sigmoid cross entropy loss function on the output unit of feed-forward network is made use of.

The first step involves carrying out the supervised pre-training of the Spectral Segregator and Order Extraction modules. The learning rate is 2.5e-04 and we train each of the modules for 60 epochs. Consequently, the parameters of the modules are fixed and the final feed-forward network is trained on the question answering task for 50 epochs with a learning rate of 2.5e-04. Finally, the learning rate is lowered to 2.5e-05 and the entire architecture  is trained (along with the modules) end-to-end for 50 epochs. We select the model with the best performance on the validation set.

\subsection{Results}

Table \ref{tab:tab1} compares the performances of CNN + LSTM, Relation Networks, FigureNet and a human baseline. These numbers are obtained on a subset of the test set(as reported by \cite{kahou2017figureqa}). The CNN + LSTM baseline is a simple architecture that concatenates the representation of an image after passing it through a CNN, with the representation of the text after passing it through an LSTM. This concatenated representation is passed through feed-forward layers to obtain the answer. The RN baseline is identical to that described in Section 3.2. The poor performance of the CNN+LSTM baseline signifies the difficulty of the task and also shows that traditional convolution architectures are not sufficient to handle relational reasoning.

\begin{table}[ht]
    \centering
    \caption{Accuracy of Model.}
    \begin{tabular}{|c||c|}
    \hline
        \emph{Model} & \emph{Accuracy} \\
        \hline \hline
        CNN + LSTM  & 59.94 \\
        \hline 
        RN(Baseline)  & 77.33 \\
        \hline
        Our Model & \textbf{84.29}  \\
        \hline
        Human & 93.29 \\
        \hline
    \end{tabular}
    
    \label{tab:tab1}
\end{table}

 Table \ref{tab:tab2} shows the performance of the models for each figure type. It can be seen that our model outperforms the baselines on all three figure types. We find that our model performs particularly well on pie charts and the performance on this figure type is closest to human performance. 
\begin{table}[h!]
    \centering
    \caption{Accuracy per figure type.}
    \resizebox{\linewidth}{!}{
    \begin{tabular}{|c||c|c|c|c|}
    \hline
        \emph{Figure Type} & \emph{CNN + LSTM} & \emph{RN(Baseline)} & \emph{Our Model} & \emph{Human} \\
        \hline \hline
        Vertical Bar & 60.84 &77.53&\textbf{87.09}& 95.90\\
        \hline 
        Horizontal Bar & 61.06&75.76&\textbf{82.19}& 96.03\\
        \hline
        Pie Chart &57.91 &78.71&\textbf{83.69}& 88.26 \\
        \hline
    \end{tabular}}
    \label{tab:tab2}
\end{table}

We observed that model performance is close to human performance for questions on maximum, minimum and comparison of plot elements. From Table \ref{tab:tab3}, we can see that questions on low median and high median are the most difficult for models as well as humans.

\begin{table}[h!]
    \centering
    \caption{Accuracy per question type.}
    \resizebox{\linewidth}{!}{
    \begin{tabular}{|c||c|c|c|c|}
    \hline
        \emph{Template} & \emph{CNN + LSTM} & \emph{RN(Baseline)} & \emph{Our Model} & \emph{Human} \\
        \hline \hline
        Is X the minimum? &60.12 &75.55&\textbf{89.86}& 97.06\\
        \hline 
        Is X the maximum? &64.70&89.29&\textbf{90.25}& 97.18\\
        \hline
        Is X the low median? &54.87&68.94&\textbf{73.74}& 86.39 \\
        \hline
        Is X the high median? &55.83&69.37&\textbf{73.71}& 86.91\\
        \hline 
        Is X less than Y? &62.31&80.63&\textbf{89.40}& 96.15\\
        \hline
        Is X greater than Y? &62.07&80.85&\textbf{89.58}& 96.15 \\
        \hline
    \end{tabular}}
    \label{tab:tab3}
\end{table}

We also show the performance of the modules in our model. We report the accuracy of the individual modules used for identifying plot elements and their relative orders. 

\begin{table}[h!]
    \centering
    \caption{Performance of the modules in the model.}
    \begin{tabular}{|c||c|}
    \hline
        \emph{Module} & \emph{Accuracy} \\
        \hline \hline
        Spectral Segregator & 80.82 \\
        \hline 
        Order Extraction  & 74.31 \\
        \hline
       
    \end{tabular}
    \label{tab:module}
\end{table}

\subsection{Ablation Studies}
We perform an ablative analysis to highlight the essentiality of different components of the model.

\subsubsection{Effect of using LSTM}
Both the Spectral Segregator and Order Extraction modules use two layered LSTM to output the identities and relative orders of the plot elements. We investigate the effect of using the LSTM network. 

First, we study the advantage of not using sampling. In the two layered LSTMs present in each of the modules, the output probabilities at time step $t-1$ are given as input to time step $t$. This is a modification to the standard approach where a vector is sampled from the output probabilities of time step $t-1$, and the sampled one-hot vector is fed as input at time step $t$. The disadvantage with the standard approach is the discrepancy during the training and testing phases. Instead, we directly feed the output probabilities of the previous time step as input to current time step. From Table \ref{tab:lstm}, it is evident that there is a drop in performance when using a sampling based approach.

\begin{table}[h!]
    \centering
    \caption{Comparing effect of LSTM modifications.}
    \begin{tabular}{|c||c|}
        \hline
            \emph{Model} & \emph{Accuracy} \\
            \hline \hline
            Our Model & 84.29 \\
            \hline 
            With sampling  &  81.61\\
            \hline
            1 layer LSTM  & 75.29 \\
            \hline
            Without LSTM
             & 73.19\\

            \hline
        \end{tabular}
    \label{tab:lstm}
\end{table}

Next, we investigate the effect of using two layers in the LSTM. We train another model that uses a single layer LSTM. We observe a huge drop in accuracy as shown in Table \ref{tab:lstm}, which signifies the greater representational capacity of a two layered LSTM. The drop in performance of the Order Extraction module was much higher than that of the Spectral Segregator module, thereby emphasizing that the second layer of the LSTM is essential for the sorting sub-task.

Finally, we train our model by removing the LSTM and predicting the 11 outputs in the modules at one go. We observed a drop in performance compared to using an LSTM. We notice that in the absence of LSTMs, the model predicts the same plot element multiple times. We hypothesize that the lack of LSTMs result in a model that is unaware of the previously predicted plot elements, resulting in repeated predictions. 

\subsubsection{Effect of using Depthwise convolutions}

Finally, we study how depthwise convolutions are essential to each module. We train our modules by replacing the depthwise convolutions with typical $3\times3$ convolutions. As shown in Table \ref{tab:depth}, the performance of the modules dropped on removing the depthwise convolutions. We hypothesize that depthwise convolutions equip the model with an ability to differentiate colours more easily.  This claim is strengthened by the fact that the Spectral Segregator module has a larger drop in performance on removing the depthwise convolutions. The ability to distinguish colours is easier when the channel information is not entirely aggregated. Since traditional convolutions sum along all channels, the learned network parameters are incapable of retaining critical information across layers, hence leading to poor performances.

\begin{table}[h]
    \centering
    \caption{Performance without Depthwise Convolutions.}
    \begin{tabular}{|c||c|c|}
        \hline
            \emph{Model} & \emph{Spectral Seg.} & \emph{Order Extraction} \\
            \hline \hline
            With Depthwise Convolutions & 80.82 & 74.31\\
            \hline 
            Without Depthwise Convolutions & 15.76 & 54.04 \\
            \hline
        \end{tabular}
    \label{tab:depth}
\end{table}

\subsection{Training time comparison}

The Relation Networks baseline has a considerably larger training time than the FigureNet architecture. The computational complexity of RNs arise from the need to process $\binom{N}{2}$ combinations of the vectors in the last CNN feature map. Each of these combinations have to pass through a MLP before they can be aggregated. The large improvement in training time can be attributed to the knowledge imbibed by the pre-training tasks.  

The Relation Networks baseline was trained on FigureQA using an open-source implementation\footnote{\href{https://github.com/vmichals/FigureQA-baseline}{https://github.com/vmichals/FigureQA-baseline}} of Relation Networks. The model was trained on a machine with a single Nvidia-1080Ti GPU and 8 CPU cores. The model was run for 600,000 steps (as done in \cite{kahou2017figureqa}).

The training for the FigureNet model was done on a machine with 4 CPU cores and a single Nvidia K80 GPU. Since the FigureNet architecture incorporates pre-training, the reported training time corresponds to the summation of the individual tasks/steps. The training times are presented in Table \ref{tab:tab7}. Note that although there exists a discrepancy in the hardware used, the configuration used to train the Relation Network is better and hence the difference in training times is expected to be larger.

\begin{table}[h!]
    \centering
    \caption{Comparing training times of RNs and FigureNet.}
    \begin{tabular}{|l||l|}
        \hline
            \emph{Model} & \emph{Time(hours)} \\
            \hline \hline
            Relation Networks & 354.79  \\
            \hline 
            FigureNet & 28.50 \\
            \hspace{0.20cm} Spectral Segregator  & \hspace{0.20cm} 7.10 \\
            \hspace{0.20cm} Order Extraction & \hspace{0.20cm} 6.58 \\
            \hspace{0.20cm} Feed-forward layer & \hspace{0.20cm} 4.25 \\
            \hspace{0.20cm} End-to End & \hspace{0.20cm} 10.47 \\
            \hline
        \end{tabular}
    \label{tab:tab7}
\end{table}

\section{Extending to beyond Synthetic Figures}

Real life scientific figures need not have a mapping between the plot element colour and name, since the plot elements can be indistinguishably coloured in each figure. Hence, there is a need to identify the plot element names from the axis/legend in the figure. Here, we give an approach for extending the current modules to real life figures:
\begin{enumerate}
    \item The bounding box annotations, as shown in Figure \ref{fig:2}, can be used to train a detection model. This model detects the bounding boxes around the plot element names on the axis or legend.
    \item Optical Character Recognition(OCR) can be used to get the plot element names from the detected bounding boxes. The detection model + OCR replaces the Spectral Segregator module that we used earlier.
    \item The Order Extraction module can be used as is, to obtain the relative ordering of plot elements.
    \item The figure representation is formed by concatenating the word embeddings of plot element names obtained, with the outputs from Order Extraction module.
    \item This figure representation, combined with the question encoding, can be used for the final question answering task on real-life scientific plots/figures.
\end{enumerate}

\section{Conclusion and Future work}
In this work, we proposed a novel architecture for question answering on categorical plots like bar graphs and pie charts. The model aims to tackle the  visual and numeric reasoning tasks by using modular components. We formulated supervised pre-training tasks to train simpler modules and then combined these modules to solve the question answering task. We ensure that each of the modules is differentiable so that once we incorporate the pre-trained modules into our network, the entire architecture can be trained end-to-end. 

Our model performs significantly better than the state-of-the-art Relation Networks baseline and the CNN+LSTM baseline. We show improvements in accuracy for each figure type and question type bridging the gap towards human-level performance. We also obtain significant improvements in training time as our model has training time that is over an order of magnitude lesser than that of Relation Networks.

In future work, we intend to improve the performance on low-median and high-median questions. Another more ambitious extension is to tackle a larger variety of question-answering tasks on real life scientific figures. This would include looking at plots which require understanding the legend and axis labels. Another line of work includes making the current model colour agnostic in order to test the model on unseen plot colour combinations.

\bibliographystyle{IEEEtran}
\bibliography{ieee.bib}

\end{document}